\documentclass{article}

\usepackage{subcaption} 
\usepackage{PRIMEarxiv}
\usepackage{xcolor}
\usepackage{amsmath}
\usepackage{natbib}
\usepackage{pgfplots}
\usepackage{subcaption}
\usepackage{tikz}
\usepackage{float}
\usepackage{placeins} 
\usepgfplotslibrary{groupplots} 
\usepackage{CJKutf8}
\usepackage[utf8]{inputenc} 
\usepackage[T1]{fontenc}    
\usepackage{hyperref}       
\usepackage{url}            
\usepackage{booktabs}       
\usepackage{amsfonts}       
\usepackage{nicefrac}       
\usepackage{microtype}      
\usepackage{lipsum}
\usepackage{fancyhdr}       
\usepackage{graphicx}       
\graphicspath{{media/}}     

\pgfplotsset{compat=1.18}
\title{Pairwise Preference Reward and Group-Based Diversity Enhancement for Superior Open-Ended Generation
}

\author{
Guining Cao$^{1,2}$\thanks{Equal contribution.}\quad
Jiaxin Peng$^{1}$\footnotemark[1]\quad
Chu Zeng$^{1,3}$\footnotemark[1]\quad
Yu Zhao$^{1}$\quad
Shuangyong Song$^{1}$\quad
Yongxiang Li$^{1}$\thanks{Corresponding author.} \\
$^{1}$Xingchen AGI Lab, China Telecom Artificial Intelligence Technology (Beijing) Co., Ltd  \hspace{2em} \\
$^{2}$School of Software and Microelectronics, Peking University   \hspace{1em} 
$^{3}$Tsinghua University \\
\texttt{riu@stu.pku.edu.cn,\hspace{1em}hygs12345@163.com,} \\
\texttt{\{pengjx3,zhaoy11,songshy,liyx25\}@chinatelecom.cn}
}

\begin{document}
\maketitle

\begin{abstract}
Current reinforcement learning(RL) methods are broadly applicable and powerful in verifiable settings where scalar rewards can be provided. However, in open-ended generation tasks, verifying the correctness of responses remains challenging, and training reward models incurs substantial computational and annotation costs. Moreover, reinforcement learning (RLVR) often leads to diversity collapse and produces stereotypical or rigid outputs — outcomes that are particularly undesirable in open-domain scenarios. We propose \textbf{Pairwise Preference Reward and Group-based Diversity Enhancement (PPR-GDE)}, a RL method that is more suitable for open-ended generation. PPR-GDE does not require scalar rewards and incorporates group-level diversity into the reward signal, it preserves the comparative structure of subjective evaluation through a pairwise preference reward, mitigates judge position bias via repeated comparisons with swapped response order, and introduces a group-based diversity reward that explicitly encourages semantic dispersion within a response group, all of these reward signals are integrated into a unified group-relative policy optimization objective. We instantiate PPR-GDE on role-playing task, experiments show that PPR-GDE achieves a better alignment quality as well as expressive diversity than strong RL baselines. Further analysis shows that pairwise preference is critical for preference alignment in subjective perspective, while the diversity metric plays an essential role in achieving superior expressive diversity and broader semantic coverage.
\end{abstract}


\section{Introduction}
Large language models (LLMs) have achieved impressive performance across diverse generation tasks, and reinforcement learning \citep{christiano2017deep,ouyang2022training} is widely used in reasoning tasks in verifiable settings such as math and coding\citep{shao2024deepseekmath}. However, aligning model with human preferences remains particularly challenging for \textit{open-ended generation tasks}, where multiple responses differ along nuanced dimensions such as style, creativity, aesthetics, persona consistency, and emotional tone. In such settings, the quality of the response is typically determined by \textit{relative human judgments} rather than absolute correctness, making both evaluation and optimization inherently difficult \citep{xu2025unified}. 

In practice, most RL pipelines rely on scalar rewards provided by automated judges to supervise preference learning \citep{christiano2017deep,ouyang2022training}. 
Pairwise supervision is often more reliable than absolute scoring for open-ended generation tasks, since it better captures subtle preference differences while reducing evaluator burden. 
Nevertheless, most existing methods eventually reduce these pairwise judgments to scalar rewards assigned to individual responses and optimize them with standard reinforcement learning algorithms such as PPO \citep{ouyang2022training,schulman2017ppo} and GRPO\citep{shao2024deepseekmath}. 
This design discards the \textit{relative structure} of subjective evaluation and introduces a mismatch between how preferences are assessed (comparatively) and how policies are optimized (independently), which can lead to unstable learning signals and suboptimal alignment. 
Recent preference-alignment methods such as DPO and P3O have highlighted the importance of operating more directly on comparative feedback \citep{rafailov2023direct,wu2023pairwise}, yet preserving pairwise preference structure within an online group-based reinforcement learning framework remains underexplored.

\begin{figure}[h]
    \centering
    \includegraphics[width=0.998\linewidth, height=7cm]{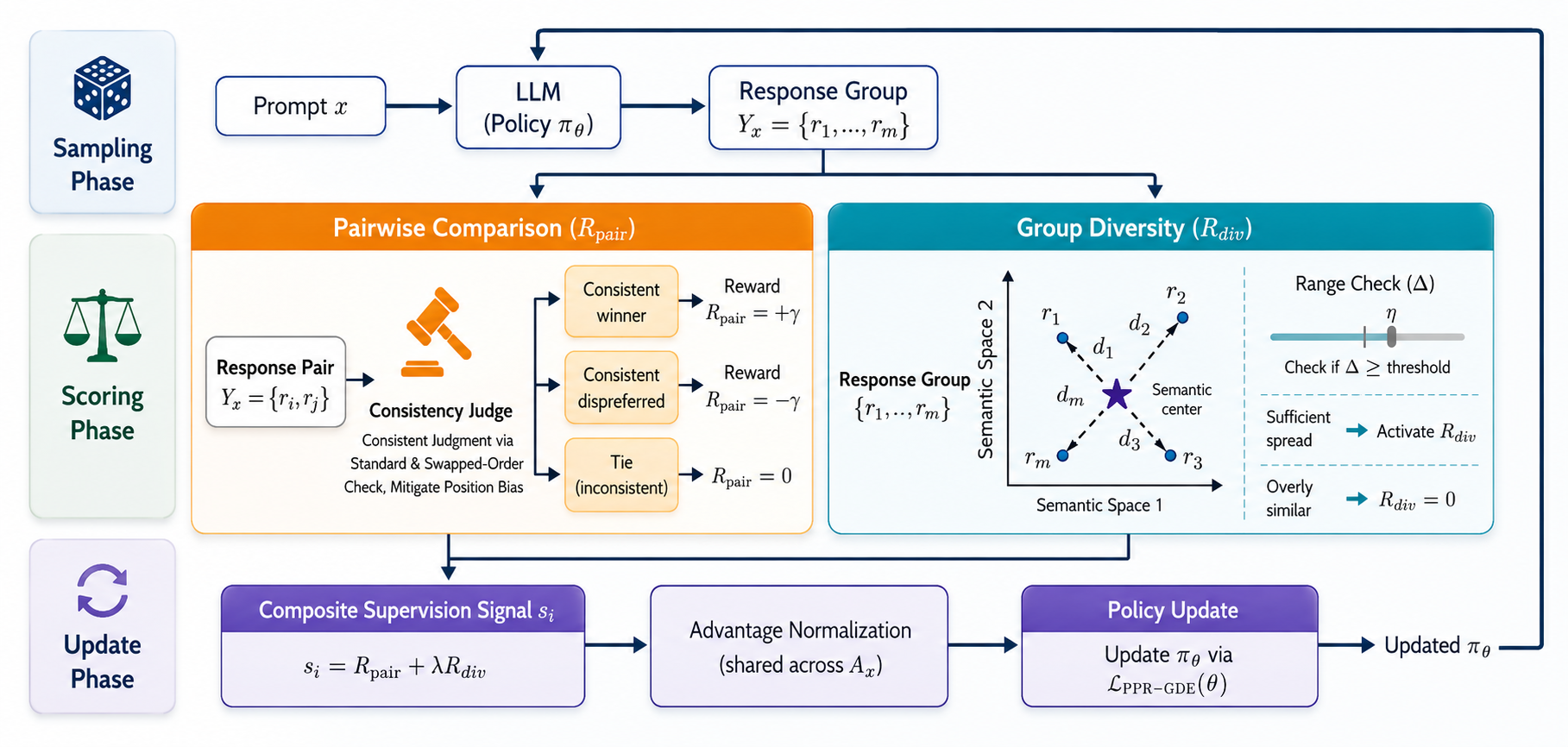}
    \caption{Overview of the proposed PPR-GDE RL method. One group of responses are generated for the  prompt and optimized jointly through pairwise preference rewards and group-based diversity enhancement, enabling stable reinforcement learning for open-ended generation tasks.}
    \label{fig:framework}
\end{figure}


The second challenge is the \textit{entropy collapse} or \textit{diversity collapse} commonly observed during RL. 
Recent studies show that policy entropy often decreases rapidly during training, causing the model to concentrate probability mass on a narrow set of already high-probability outputs and weakening exploratory coverage \citep{ahmed2019understanding,wen2024entropy,cui2025entropy,cheng2025reasoning}. 
This problem is especially harmful for open-ended generation tasks, where a well-aligned model should not only deliver high-quality responses, but also maintain response diversity under repeated inputs of an identical prompt to avoid stereotyped outputs.
Existing diversity-preserving strategies mainly regularize token- or policy-level entropy, or promote lexical diversity within a single response \citep{li-etal-2016-diversity,wen2024entropy,cui2025entropy}. 
However, for open-ended generation, a stronger notion of diversity is required: \textit{semantic dispersion across multiple responses generated for the same prompt}. 
This is particularly important for role-playing dialogue \citep{wang-etal-2024-rolellm,tu-etal-2024-charactereval}, where natural interaction requires not only persona fidelity, but also expressive flexibility---a good role-playing model should be able to produce diverse yet still character-consistent responses to the same input.

To address these challenges, we propose \textbf{Pairwise Preference Reward and Group-based Diversity Enhancement (PPR-GDE)}, a reinforcement learning method tailored for open-ended generation tasks. 
Within each group of responses generated by model, we preserve the comparative nature of subjective evaluation through a \textit{pairwise preference reward} formulation, allowing the policy to learn from relative quality distinctions rather than relying on potentially noisy absolute reward magnitudes. 
Moreover, our pairwise design can mitigate position bias by performing repeated comparisons with swapped response order \citep{shi2024judging}. 
Concurrently, we introduce a \textit{group-based diversity} mechanism that explicitly encourages semantic dispersion among responses within the group, thereby preventing diversity collapse during reinforcement learning. 
Consequently, PPR-GDE not only improves alignment quality but also achieves superior expressive diversity and coverage. 
While we instantiate this framework in the role-playing domain, its formulation is broadly applicable to various open-ended generation tasks.
Figure~\ref{fig:framework} illustrates the overall framework of PPR-GDE.


\noindent\textbf{The contributions of this study are summarized as follows:}
\begin{itemize}

    \item We propose \textbf{P}airwise \textbf{P}reference \textbf{R}eward and \textbf{G}roup-based \textbf{D}iversity \textbf{E}nhancement \textbf{(PPR-GDE)}, which use pairwise preference signal rather than scalar reward and novelly unities group-based semantic diversity as the optimization objection.

    \item We formulate PPR-GDE as a reinforcement learning framework for open-ended generation tasks and instantiate it on the role-playing scenario as a representative case, achieving improved diversity and alignment quality while maintaining stable reinforcement learning dynamics relative to PPO and GRPO.

    \item We construct our role-playing evaluation bench based on  RoleBench and mCSQA, the experimental results and analyses demonstrate that PPR-GDE maintains higher entropy and slower collapse than GRPO, within the generated response groups, PPR-GDE can yield 30\% more clusters on average.


\end{itemize}

\section{Preliminary}

\subsection{Reinforcement Learning for Language Models}

We consider language generation as a sequential decision-making process, where a language model acts as a policy that generates tokens conditioned on a prompt. 
Let $x \sim \mathcal{D}$ denote a prompt sampled from a data distribution, and let $\pi_{\theta}$ denote a parameterized policy. 
The model generates a response $r = (y_1, y_2, \ldots, y_T)$ according to

\begin{equation}
r \sim \pi_{\theta}(\cdot \mid x), \quad 
\pi_{\theta}(r \mid x) = \prod_{t=1}^{T} \pi_{\theta}(y_t \mid x, y_{<t}).
\end{equation}

Reinforcement learning aims to optimize the policy parameters $\theta$ such that the expected reward over generated responses is maximized:

\begin{equation}
\max_{\theta} \; 
\mathbb{E}_{x \sim \mathcal{D},\, r \sim \pi_{\theta}(\cdot \mid x)} 
\left[ R(x, r) \right],
\end{equation}

where $R(x,r)$ denotes a scalar reward that evaluates the quality of the generated response.

In practice, policy-gradient methods such as Proximal Policy Optimization (PPO) are commonly used to optimize this objective \citep{schulman2017ppo}.  
Let $\pi_{\theta_{\text{old}}}$ denote the behavior policy used for sampling. 
The policy update can be written in the following importance-weighted form:

\begin{equation}
\mathcal{L}(\theta) =
\mathbb{E}\left[
\frac{\pi_{\theta}(r \mid x)}{\pi_{\theta_{\text{old}}}(r \mid x)} A(x,r)
\right],
\end{equation}

where $A(x,r)$ denotes an advantage function estimating the relative quality of the sampled response. 
In language model alignment, the underlying reward signal is often derived from human feedback or preference judgments \citep{christiano2017deep,ouyang2022training}.

\subsection{Group Relative Policy Optimization}

Compared to PPO, GRPO eliminates the value function and estimates the advantage in a group-relative
manner\citep{yu2025dapo}.

Given a prompt $x$, the behavior policy $\pi_{\theta_{\text{old}}}$ samples a group of responses:

\begin{equation}
\mathcal{Y}_x = \{ r_1, r_2, \ldots, r_G \}, \quad r_i \sim \pi_{\theta_{\text{old}}}(\cdot \mid x).
\end{equation}

Each response receives a reward $R_i$, and the group-relative advantage is computed by normalizing rewards within the group:

\begin{equation}
\hat{A}_{i} =
\frac{R_i - \mathrm{mean}(\{R_j\}_{j=1}^{G})}
{\mathrm{std}(\{R_j\}_{j=1}^{G})}.
\end{equation}

Following common GRPO-style notation, this group-level advantage is shared across the tokens of the same response and is written as $\hat{A}_{i,t}$ in the policy objective. 
This group-relative normalization removes the need for an explicit value function and encourages the policy to improve responses relative to alternative candidates generated under the same prompt \citep{shao2024deepseekmath}.

The GRPO optimization objective follows a clipped policy-gradient formulation with an additional KL regularization term:
\begin{equation}
\mathcal{L}_{\text{GRPO}}(\theta) =
\mathbb{E}\left[
\frac{1}{G}
\sum_{i=1}^{G}
\frac{1}{|r_i|}
\sum_{t=1}^{|r_i|}
\left(
\min
\left(
\rho_{i,t}(\theta)\hat{A}_{i,t},
\operatorname{clip}\big(\rho_{i,t}(\theta), 1-\epsilon, 1+\epsilon\big)\hat{A}_{i,t}
\right)
-
\beta D_{\mathrm{KL}}\!\left(\pi_{\theta}\,\|\,\pi_{\mathrm{ref}}\right)
\right)
\right],
\end{equation}

where
\begin{equation}
\rho_{i,t}(\theta) =
\frac{
\pi_{\theta}(y_{i,t} \mid x, y_{i,<t})
}{
\pi_{\theta_{\text{old}}}(y_{i,t} \mid x, y_{i,<t})
}.
\end{equation}

Here, $\epsilon$ denotes the clipping parameter, $\beta$ is the KL regularization coefficient, and $\pi_{\mathrm{ref}}$ is the reference policy \citep{shao2024deepseekmath}. 
The KL term regularizes the updated policy toward the reference model, thereby improving training stability and preventing excessive policy drift.

\section{Method}

PPR-GDE method incorporates two key components: (1) a \textit{pairwise preference reward} that learns from relative quality comparisons, which aligns with subjective evaluation which differs in style, aesthetics, structure, or semantic realization; and (2) a \textit{group-based diversity} mechanism that fosters semantic dispersion to prevent diversity collapse.

\subsection{Problem Setup and Framework Overview}

Let $x \sim \mathcal{D}$ denote a prompt sampled from the training distribution. 
At each policy update, the behavior policy $\pi_{\theta_{\mathrm{old}}}$ samples $G$ candidate responses
\begin{equation}
r_i = (y_{i,1}, y_{i,2}, \ldots, y_{i,|r_i|}) \sim \pi_{\theta_{\mathrm{old}}}(\cdot \mid x), \qquad i = 1, \dots, G,
\end{equation}
which form a prompt-specific response group
\begin{equation}
\mathcal{Y}_x = \{ r_1, r_2, \ldots, r_G \}.
\end{equation}

Unlike conventional reinforcement learning methods that score each response independently, PPR-GDE evaluates responses within the group. 
This design better matches open-ended generation settings, where the relative quality of two candidate responses is often easier to assess than their absolute quality in isolation.

For each response $r_i$, we define a composite supervision signal
\begin{equation}
{s_i = R_i^{\mathrm{pair}} + \lambda R_i^{\mathrm{div}}},
\label{eq:composite_signal}
\end{equation}
where $R_i^{\mathrm{pair}}$ is the pairwise preference reward, $R_i^{\mathrm{div}}$ is the diversity reward, and $\lambda$ controls the contribution of diversity. 
The resulting signal is then normalized within the valid response subset to construct a group-relative advantage for policy optimization.

Overall, the training pipeline proceeds as follows. 
Given a prompt, the policy first samples a response group $\mathcal{Y}_x$. 
Pairwise comparison and group-based diversity evaluation are then performed to obtain the supervision signal for each response. 
Finally, the policy is updated using a GRPO-style clipped objective \citep{shao2024deepseekmath}.

\subsection{Pairwise Preference Reward}

For many open-ended generation tasks, especially subjective tasks such as role-playing, absolute quality scores are often difficult to define reliably. 
In contrast, pairwise comparison provides a more stable and interpretable supervision signal by directly asking which of two responses better satisfies the task requirement \citep{christiano2017deep,ouyang2022training,rafailov2023dpo}. 
Based on this observation, we construct response-level preference rewards from pairwise judgments within each sampled response group.


To mitigate position bias in the preference judge, we compare each response pair twice, with the order of the two responses swapped between the two judge inputs \citep{shi-etal-2025-judging}. 
Specifically, the response group $\mathcal{Y}_x$ is randomly partitioned into disjoint pairs. 
For a response $r_i$, let $r_{p(i)}$ denote its paired counterpart, and let
\[
\phi(x,a,b)\in\{-1,1\}
\]
denote the judge output for prompt $x$, where $\phi(x,a,b)=1$ means that the judge prefers the first response $a$ over the second response $b$, and $\phi(x,a,b)=-1$ means the opposite. 
Since the judge only outputs a strict preference, tie is not produced directly by the judge.

We define the response-level preference label
\begin{equation}
z_i =
\begin{cases}
1, & \text{if } \phi(x,r_i,r_{p(i)}) = 1 \text{ and } \phi(x,r_{p(i)},r_i) = -1,\\
-1, & \text{if } \phi(x,r_i,r_{p(i)}) = -1 \text{ and } \phi(x,r_{p(i)},r_i) = 1,\\
0, & \text{otherwise},
\end{cases}
\label{eq:pair_label}
\end{equation}
where $z_i = 1$ indicates that $r_i$ is consistently preferred over its paired counterpart across both input orders, $z_i = -1$ indicates that it is consistently dispreferred, and $z_i = 0$ denotes a tie induced by inconsistent judgments across the two order-swapped comparisons. 
For the paired response $r_{p(i)}$, we correspondingly have $z_{p(i)} = -z_i$.

We then define the active-response set as
\begin{equation}
{\mathcal{A}_x = \{\, r_i \in \mathcal{Y}_x \mid z_i \neq 0 \,\}, \qquad N_x = |\mathcal{A}_x|.}
\label{eq:active_set}
\end{equation}
Only responses in $\mathcal{A}_x$ contribute to training. 
If $N_x = 0$, the prompt provides no valid optimization signal in the current update.

For each active response $r_i \in \mathcal{A}_x$, the pairwise preference reward is defined as
\begin{equation}
R_i^{\mathrm{pair}} =
\begin{cases}
\gamma \alpha_i, & \text{if } z_i = 1,\\
-\gamma, & \text{if } z_i = -1,
\end{cases}
\label{eq:pair_reward}
\end{equation}
where $\gamma > 0$ is a base reward constant. 
No reward is assigned when $z_i = 0$, since tied responses are removed from optimization rather than treated as zero-reward samples.

To reduce verbosity bias, we further apply length normalization to winning responses. 
Let $L_i$ denote the length of response $r_i$, and let $\mathcal{W}_x = \{\, r_i \in \mathcal{A}_x \mid z_i = 1 \,\}$ denote the set of winning responses for prompt $x$. 
The length-normalization coefficient is defined as
\begin{equation}
\alpha_i =
\frac{
\min_{r_j \in \mathcal{W}_x} L_j
}{
L_i
}.
\label{eq:length_norm}
\end{equation}
This formulation assigns the largest reward to the shortest winning response while softly penalizing unnecessarily long outputs.

\subsection{Group-based Diversity Reward}

Optimizing only for pairwise preference may cause the policy to concentrate on a narrow set of high-scoring response patterns, thereby reducing semantic diversity among generated outputs. 
To mitigate this issue, we introduce a group-based diversity reward that encourages responses to explore different semantic regions.

Given a prompt $x$, the sampled responses $\mathcal{Y}_x$ are partitioned into disjoint subgroups of size $M$ for diversity computation. 
For each response $r_i$, let $\mathcal{G}_x(i)$ denote the subgroup containing $r_i$. 
We first obtain its semantic embedding $e_i = f(r_i)$ using a sentence embedding model $f(\cdot)$, and then measure how far it deviates from the semantic center of its subgroup:
\begin{equation}
d_i =
\left\|
e_i -
\frac{1}{|\mathcal{G}_x(i)|}
\sum_{r_j \in \mathcal{G}_x(i)} e_j
\right\|_2.
\label{eq:div_distance}
\end{equation}

Let 
\[
\Delta(\mathcal{G}_x(i)) = \max_{r_j \in \mathcal{G}_x(i)} d_j - \min_{r_j \in \mathcal{G}_x(i)} d_j
\]
denote the diversity range of the subgroup containing $r_i$. 
The diversity reward for response $r_i$ is then defined as
\begin{equation}
R_i^{\mathrm{div}} =
\begin{cases}
0, & \text{if } \Delta(\mathcal{G}_x(i)) < \eta, \\[6pt]
\dfrac{
d_i - \min_{r_j \in \mathcal{G}_x(i)} d_j
}{
\max_{r_j \in \mathcal{G}_x(i)} d_j - \min_{r_j \in \mathcal{G}_x(i)} d_j
}, & \text{otherwise},
\end{cases}
\label{eq:div_reward}
\end{equation}
where $\eta > 0$ is a threshold controlling whether the subgroup exhibits sufficient semantic dispersion to activate diversity supervision.

In this way, diversity reward is provided only when a subgroup shows meaningful semantic spread. 
When the responses in a subgroup are overly similar, the diversity signal is suppressed by setting $R_i^{\mathrm{div}} = 0$ for all responses in that subgroup. 
Notably, this computation is independent of pairwise preference optimization; tied samples are discarded only in the final policy update stage.

\subsection{PPR-GDE Optimization Objective}

Based on Eq.~\eqref{eq:composite_signal}, we normalize the composite supervision signal over the active response subset $\mathcal{A}_x$ to obtain the response-level advantage:
\begin{equation}
{\hat{A}_i} =
\frac{
{s_i} - \operatorname{mean}\!\left(\{{s_j}\}_{r_j \in {\mathcal{A}_x}}\right)
}{
\operatorname{std}\!\left(\{{s_j}\}_{r_j \in {\mathcal{A}_x}}\right)
}.
\label{eq:ppr_gde_adv}
\end{equation}
Following the GRPO-style token-level notation, this response-level advantage is shared across all tokens of the same response, i.e., $\hat{A}_{i,t} = \hat{A}_i$ for all $t$.

Using this advantage, PPR-GDE adopts the same clipped policy-gradient form as GRPO:
\begin{equation}
\mathcal{L}_{\mathrm{PPR\text{-}GDE}}(\theta)
=
\mathbb{E}_{x \sim \mathcal{D},\, {N_x > 0}}
\left[
\frac{1}{{N_x}}
\sum_{r_i \in {\mathcal{A}_x}}
\frac{1}{|r_i|}
\sum_{t=1}^{|r_i|}
\left(
\min\left(
\rho_{i,t}(\theta){\hat{A}_{i,t}},
\operatorname{clip}\big(\rho_{i,t}(\theta), 1-\epsilon, 1+\epsilon\big){\hat{A}_{i,t}}
\right)
-
\beta D_{\mathrm{KL}}\!\left(\pi_\theta \,\|\, \pi_{\mathrm{ref}}\right)
\right)
\right],
\label{eq:ppr_gde_obj}
\end{equation}
with
\begin{equation}
\rho_{i,t}(\theta) =
\frac{
\pi_\theta(y_{i,t} \mid x, y_{i,<t})
}{
\pi_{\theta_{\mathrm{old}}}(y_{i,t} \mid x, y_{i,<t})
}.
\label{eq:ppr_gde_ratio}
\end{equation}

Here, $\epsilon$ is the clipping parameter, $\beta$ is the KL regularization coefficient, and $\pi_{\mathrm{ref}}$ is the reference policy. 
Only active responses in $\mathcal{A}_x$ participate in advantage normalization and policy optimization; if all pairwise comparisons for a prompt are ties, i.e., $N_x = 0$, that prompt contributes no update in the current training step.

Compared with GRPO, the optimization form remains unchanged, while the supervision signal and the valid sample set are redefined. 
Specifically, the conventional scalar reward is replaced by the composite signal $s_i$, and the normalization and averaging are performed over the active subset $\mathcal{A}_x$ with size $N_x$ rather than the full sampled group $\mathcal{Y}_x$ with size $G$. 
Therefore, the red-marked terms in the above equations indicate exactly where PPR-GDE differs from GRPO.

Overall, PPR-GDE preserves the group-relative optimization structure of GRPO while redefining the signal used for advantage estimation. 
By replacing the conventional reward with the composite supervision signal $s_i$, the proposed objective encourages the policy to generate responses that are both subjectively preferred and semantically diverse.

\section{Experiments}

We evaluate the proposed PPR-GDE framework on role-playing generation from three complementary perspectives: single-turn role-playing quality, multi-turn role consistency, and response diversity.

\subsection{Experimental Setup}

\paragraph{Base Model and Baselines.}
Experiments are conducted based on \textbf{Qwen2.5-1.5B-Instruct}, \textbf{Qwen2.5-3B-Instruct} and \textbf{Qwen3-0.6B}. 
We compare PPR-GDE against three baselines built on the same backbone: the original \textbf{Base Model} without further alignment, \textbf{PPO}, and \textbf{GRPO}.

\paragraph{Training Data.}
Our training data is constructed from two complementary sources. 
The first source is \textbf{RoleBench} \citep{wang-etal-2024-rolellm}, from which we use role-specific samples to improve the model's role-playing ability, especially its understanding of role background and character-specific knowledge. 
The second source is \textbf{mCSQA} \citep{sakai-etal-2024-mcsqa}, which is used to strengthen the model's basic response quality under role-playing settings. 
For the latter, we assign each sampled question a role profile and background, and use \textbf{Qwen3-32B} as a compatibility judge to filter out question--role pairs that conflict with the assigned character setting. 
After merging the two sources, we obtain a training set of \textbf{11,895} examples and a held-out pool of \textbf{1,024} examples, which is evenly split into validation and test sets.

\paragraph{Training Protocol.}
During training, multiple candidate responses are sampled for each prompt. 
For PPO and GRPO, policy updates are driven by scalar reward signals. 
For PPR-GDE, responses are compared pairwise to construct preference-based supervision, together with an additional group-based diversity reward. 
This design encourages the model to generate responses that are not only preferred, but also semantically diverse.

\paragraph{Hyperparameter Settings.}
To ensure fair comparison, all methods share the same backbone model, decoding setup, optimizer settings, and training budget unless otherwise specified. 
Thus, the primary difference across methods lies in the alignment objective rather than implementation details.






\subsection{Evaluation Metrics}

We evaluate all methods from three complementary perspectives: role-playing quality, multi-turn role consistency, and response diversity. 
Unless otherwise specified, higher scores indicate better performance.

\paragraph{Role-Playing Quality.}
Following \citet{wang-etal-2024-rolellm}, we report three benchmark-native metrics for single-turn role-playing quality: \textbf{CUS}, \textbf{RAW}, and \textbf{SPE}. 
These metrics measure role-style consistency, instruction-following quality, and role-specific knowledge expression, respectively. 
Among them, \textbf{CUS} is evaluated on the full test set. 
\textbf{RAW} is evaluated on the subset derived from mCSQA, and mainly reflects the model's basic knowledge and general response ability under role constraints. 
\textbf{SPE} is evaluated on the subset derived from RoleBench, and is used to assess the model's understanding of role background and character-specific knowledge \citep{wang-etal-2024-rolellm,sakai-etal-2024-mcsqa}.

\paragraph{Multi-Turn Role Consistency.}
For multi-turn evaluation, we follow \citet{tu-etal-2024-charactereval} and report \textbf{Knowledge-Exposure (KE)}, \textbf{Knowledge-Accuracy (KA)}, and \textbf{Persona-Utterance (PU)} using the official CharacterEval dataset and evaluation pipeline. 
Although CharacterEval contains a broader set of evaluation dimensions, we retain these three metrics because they are the most informative for measuring multi-turn role-playing consistency in our setting \citep{tu-etal-2024-charactereval}.

\paragraph{Response Diversity.}
Beyond role fidelity, we further evaluate response diversity on our test set from both lexical and semantic perspectives. 
We report \textbf{Distinct-2} \citep{li-etal-2016-diversity} as the lexical diversity metric. 
For semantic diversity, we use two complementary measures. 
First, we compute \textbf{Semantic Nearest-Neighbor Distance (SNND)}, defined as the average distance from each response embedding to its nearest semantic neighbor:
\[
\mathrm{SNND}
=
\frac{1}{|\mathcal{R}|}
\sum_{i=1}^{|\mathcal{R}|}
\min_{j \neq i}
\left\|
\mathbf{e}_i-\mathbf{e}_j
\right\|_2,
\]
where $\mathbf{e}_i$ denotes the embedding of response $r_i$. 
A larger SNND indicates stronger local semantic dispersion. 
We also report \textbf{Number of Clusters (NoC)}, which counts the number of semantic groups obtained by clustering the response embeddings with a fixed clustering algorithm. 
A larger NoC indicates broader global semantic coverage.

Overall, Distinct-2 measures lexical variation, while SNND and NoC characterize semantic diversity from the perspectives of local dispersion and global coverage, respectively.

\subsection{Overall Results}
\label{sec:overall_results}

Table~\ref{tab:main_results} summarizes the main results on three backbone scales: \textbf{Qwen2.5-3B-Instruct}, \textbf{Qwen2.5-1.5B-Instruct}, and \textbf{Qwen3-0.6B}. 
Overall, RL-based alignment substantially improves role-playing performance over the corresponding base model across all three scales. 
Among the aligned methods, \textbf{PPR-GDE} consistently achieves the strongest overall balance between role-playing quality, multi-turn consistency, and response diversity.

On \textbf{Qwen2.5-3B-Instruct} and \textbf{Qwen2.5-1.5B-Instruct}, PPR-GDE achieves the best average score on \textbf{Role-Playing Ability}, and also obtains the best results on \textbf{CUS} and \textbf{SPE}. 
On \textbf{Qwen3-0.6B}, PPR-GDE still achieves the best \textbf{CUS} and \textbf{SPE} scores, but its role-playing average is slightly below PPO and GRPO due to a lower \textbf{RAW} score. 
This suggests that the proposed pairwise preference reward is particularly effective for improving the subjective dimensions of role-playing alignment, especially persona consistency and role-specific knowledge expression.

PPR-GDE also achieves the best average score on \textbf{Multi-Turn Consistency} for all three backbones. 
In particular, it consistently performs strongly on \textbf{PU}, while remaining competitive on \textbf{KE} and \textbf{KA}. 
At the same time, the table reveals a consistent trade-off: methods with stronger role-playing quality and richer expressive coverage often obtain lower \textbf{RAW} scores. 
We defer a more detailed analysis of this phenomenon to Section~\ref{sec:ablation}, where the ablation study shows that diversity-oriented optimization can improve expressive coverage while partially sacrificing general response quality under role constraints.

For \textbf{response diversity}, the unaligned base model remains the most diverse, which is consistent with the general tendency of RL alignment to narrow the response distribution. 
However, among the aligned methods, PPR-GDE achieves the best diversity average on all three backbones. 
Although it does not dominate every individual diversity metric at every scale, it consistently preserves broader semantic variation than PPO and GRPO, especially in terms of \textbf{NoC} and \textbf{SNND}. 
This shows that the proposed diversity-aware reward effectively mitigates semantic collapse during RL.

\begin{table*}[htbp]
    \centering
    \small
    \caption{Main results on role-playing quality, multi-turn consistency, and response diversity across two backbone scales. Best results within each panel are shown in \textbf{bold}, and second-best results are \underline{underlined}. Higher is better for all metrics.}
    \label{tab:main_results}
    \setlength{\tabcolsep}{4pt}
\begin{tabular}{lcccccccccccc}
    \toprule
    \textbf{Method} 
    & \multicolumn{4}{c}{\textbf{Role-Playing Ability}} 
    & \multicolumn{4}{c}{\textbf{Multi-Turn Consistency}} 
    & \multicolumn{4}{c}{\textbf{Diversity}} \\
    \cmidrule(lr){2-5}\cmidrule(lr){6-9}\cmidrule(lr){10-13}
    & \textbf{CUS} & \textbf{RAW} & \textbf{SPE} & \textbf{avg.}
    & \textbf{KE} & \textbf{KA} & \textbf{PU} & \textbf{avg.}
    & \textbf{Distinct-2} & \textbf{SNND} & \textbf{NoC} & \textbf{avg.} \\
    \midrule
    \multicolumn{13}{c}{\textbf{Qwen2.5-3B-Instruct}} \\
    \midrule
    w/o RL
    & 4.04 & \textbf{3.11} & 4.20 & 3.78
    & 2.17 & 2.83 & \underline{2.95} & 2.65
    & \textbf{0.67} & \textbf{0.14} & \textbf{3.28} & \textbf{1.36} \\

    PPO
    & 4.71 & \underline{2.74} & 4.72 & \underline{4.06}
    & 2.48 & \textbf{3.10} & 2.93 & 2.84
    & 0.61 & 0.10 & 1.54 & 0.75 \\

    GRPO
    & \underline{4.88} & 2.51 & \underline{4.79} & \underline{4.06}
    & \underline{2.61} & \underline{3.05} & 2.89 & \underline{2.85}
    & \underline{0.64} & \underline{0.12} & 1.98 & 0.91 \\

    \textbf{PPR-GDE (Ours)}
    & \textbf{4.99} & 2.60 & \textbf{4.93} & \textbf{4.17}
    & \textbf{2.65} & 2.94 & \textbf{3.03} & \textbf{2.87}
    & 0.61 & \underline{0.12} & \underline{2.27} & \underline{1.00} \\

    \midrule
    \multicolumn{13}{c}{\textbf{Qwen2.5-1.5B-Instruct}} \\
    \midrule
    w/o RL
    & 3.21 & \textbf{3.00} & 3.49 & 3.23
    & 1.82 & 2.70 & 2.79 & 2.44
    & \textbf{0.68} & \textbf{0.16} & \textbf{4.08} & \textbf{1.64} \\

    PPO
    & 4.77 & 2.16 & 4.63 & 3.85
    & \underline{2.37} & \underline{2.93} & 2.68 & 2.66
    & 0.55 & 0.10 & 1.97 & 0.87 \\

    GRPO
    & \underline{4.81} & 2.44 & \underline{4.66} & \underline{3.97}
    & 2.33 & \underline{2.93} & \underline{2.85} & \underline{2.70}
    & \underline{0.63} & 0.12 & 2.39 & 1.05 \\

    \textbf{PPR-GDE (Ours)}
    & \textbf{4.94} & \underline{2.82} & \textbf{4.86} & \textbf{4.21}
    & \textbf{2.57} & \textbf{3.01} & \textbf{2.99} & \textbf{2.86}
    & 0.55 & \underline{0.14} & \underline{3.02} & \underline{1.24} \\

    \midrule
    \multicolumn{13}{c}{\textbf{Qwen3-0.6B}} \\
    \midrule
    w/o RL
    & 3.38 & 1.79 & 3.35 & 2.84
    & 2.23 & 2.68 & 2.61 & 2.51
    & \textbf{0.60} & \textbf{0.14} & \textbf{3.53} & \textbf{1.42} \\

    PPO
    & \underline{4.83} & \underline{2.62} & \underline{4.79} & \underline{4.08}
    & \underline{2.42} & \underline{2.89} & \underline{2.68} & \underline{2.66}
    & 0.47 & 0.09 & 1.60 & 0.72 \\

    GRPO
    & 4.80 & \textbf{2.99} & 4.66 & \textbf{4.15}
    & \textbf{2.45} & 2.86 & 2.64 & 2.65
    & 0.20 & 0.08 & 1.39 & 0.56 \\

    \textbf{PPR-GDE (Ours)}
    & \textbf{4.97} & 2.25 & \textbf{4.94} & 4.05
    & 2.39 & \textbf{3.06} & \textbf{3.04} & \textbf{2.83}
    & \underline{0.50} & \underline{0.12} & \underline{2.04} & \underline{0.89} \\

    \bottomrule
\end{tabular}
\end{table*}

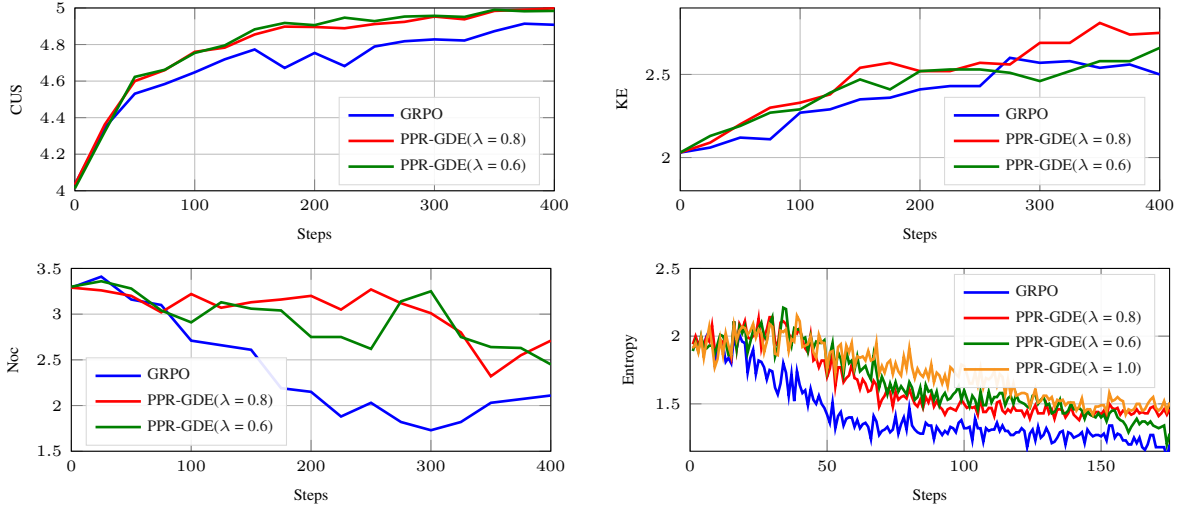
\begin{figure}[H]
\centering

\begin{subfigure}[t]{0.48\textwidth}
\centering
\begin{tikzpicture}
\begin{axis}[
    width=\linewidth,
    label style={font=\tiny},
    tick label style={font=\tiny}, 
    legend pos=south east,    
    legend style={font=\tiny, fill opacity=0.9, draw=gray!30},
    legend cell align={left}, 
    height=4.cm,
    xmin=0, xmax=400,          
    ymin=4, ymax=5, 
    xlabel={Steps},
    ylabel={CUS},
    grid=both,
]
\addplot[
    color=blue,
    line width=1pt,
] coordinates {
    (0, 4.033)
    (25, 4.346)
    (50, 4.531)
    (75, 4.584)
    (100, 4.648)
    (125, 4.719)
    (150, 4.773)
    (175, 4.672)
    (200, 4.754)
    (225, 4.682)
    (250, 4.789)
    (275, 4.818)
    (300, 4.828)
    (325, 4.822)
    (350, 4.873)
    (375, 4.914)
    (400, 4.908)
};
\addlegendentry{GRPO}
\addplot[
    color=red,
    line width=1pt,
] coordinates {
    (0, 4.033)
    (25, 4.363)
    (50, 4.6)
    (75, 4.66)
    (100, 4.76)
    (125, 4.783)
    (150, 4.855)
    (175, 4.898)
    (200, 4.896)
    (225, 4.889)
    (250, 4.912)
    (275, 4.924)
    (300, 4.953)
    (325, 4.938)
    (350, 4.984)
    (375, 4.99)
    (400, 4.998)
};
\addlegendentry{PPR-GDE($\lambda$ = 0.8)}
\addplot[
    color=green!50!black,
    line width=1pt,
] coordinates {
    (0, 4.013)
    (25, 4.324)
    (50, 4.623)
    (75, 4.662)
    (100, 4.754)
    (125, 4.795)
    (150, 4.883)
    (175, 4.918)
    (200, 4.906)
    (225, 4.947)
    (250, 4.928)
    (275, 4.953)
    (300, 4.957)
    (325, 4.951)
    (350, 4.99)
    (375, 4.982)
    (400, 4.984)
};
\addlegendentry{PPR-GDE($\lambda$ = 0.6)}
\end{axis}
\end{tikzpicture}
\label{fig:cus}
\end{subfigure}
\begin{subfigure}[t]{0.48\textwidth}
\centering
\begin{tikzpicture}
\begin{axis}[
    width=\linewidth,
    label style={font=\tiny},
    tick label style={font=\tiny}, 
    legend pos=south east,    
    legend style={font=\tiny, fill opacity=0.9, draw=gray!30},
    legend cell align={left}, 
    height=4.cm,
    xmin=0, xmax=400,          
    ymin=1.8, ymax=2.9, 
    xlabel={Steps},
    ylabel={KE},
    grid=both,
]
\addplot[
    color=blue,
    line width=1pt,
] coordinates {
    (0, 2.03)
    (25, 2.06)
    (50, 2.12)
    (75, 2.11)
    (100, 2.27)
    (125, 2.29)
    (150, 2.35)
    (175, 2.36)
    (200, 2.41)
    (225, 2.43)
    (250, 2.43)
    (275, 2.6)
    (300, 2.57)
    (325, 2.58)
    (350, 2.54)
    (375, 2.56)
    (400, 2.5)
};
\addlegendentry{GRPO}
\addplot[
    color=red,
    line width=1pt,
] coordinates {
    (0, 2.03)
    (25, 2.09)
    (50, 2.20)
    (75, 2.3)
    (100, 2.33)
    (125, 2.38)
    (150, 2.54)
    (175, 2.57)
    (200, 2.52)
    (225, 2.52)
    (250, 2.57)
    (275, 2.56)
    (300, 2.69)
    (325, 2.69)
    (350, 2.81)
    (375, 2.74)
    (400, 2.75)
};
\addlegendentry{PPR-GDE($\lambda$ = 0.8)}
\addplot[
    color=green!50!black,
    line width=1pt,
] coordinates {
    (0, 2.03)
    (25, 2.13 )
    (50, 2.19)
    (75, 2.27)
    (100, 2.29)
    (125, 2.39)
    (150, 2.47)
    (175, 2.41)
    (200, 2.52)
    (225, 2.53)
    (250, 2.53)
    (275, 2.51)
    (300, 2.46)
    (325, 2.52)
    (350, 2.58)
    (375, 2.58)
    (400, 2.66)
};
\addlegendentry{PPR-GDE($\lambda$ = 0.6)}
\end{axis}
\end{tikzpicture}
\label{fig:ke}
\end{subfigure}
\begin{subfigure}[t]{0.48\textwidth}
\centering
\begin{tikzpicture}
\begin{axis}[
    width=\linewidth,
    label style={font=\tiny},
    tick label style={font=\tiny}, 
    legend pos=south west,    
    legend style={font=\tiny, fill opacity=0.9, draw=gray!30, },
    legend cell align={left}, 
    height=4.cm,
    xmin=0, xmax=400,          
    ymin=1.5, ymax=3.5, 
    xlabel={Steps},
    ylabel={Noc},
    grid=both,
]
\addplot[
    color=blue,
    line width=1pt,
] coordinates {
    (0, 3.29)
    (25, 3.41)
    (50, 3.16)
    (75, 3.1)
    (100, 2.71)
    (125, 2.66)
    (150, 2.61)
    (175, 2.19)
    (200, 2.15)
    (225, 1.88)
    (250, 2.03)
    (275, 1.82)
    (300, 1.73)
    (325, 1.82)
    (350, 2.03)
    (375, 2.07)
    (400, 2.11)
};
\addlegendentry{GRPO}
\addplot[
    color=red,
    line width=1pt,
] coordinates {
    (0, 3.29)
    (25, 3.26)
    (50, 3.2)
    (75, 3.02)
    (100, 3.22)
    (125, 3.07)
    (150, 3.13)
    (175, 3.16)
    (200, 3.2)
    (225, 3.05)
    (250, 3.27)
    (275, 3.12)
    (300, 3.01)
    (325, 2.8)
    (350, 2.32)
    (375, 2.55)
    (400, 2.71)
};
\addlegendentry{PPR-GDE($\lambda$ = 0.8)}
\addplot[
    color=green!50!black,
    line width=1pt,
] coordinates {
    (0, 3.3)
    (25, 3.36)
    (50, 3.28)
    (75, 3.04)
    (100, 2.91)
    (125, 3.13)
    (150, 3.06)
    (175, 3.04)
    (200, 2.75)
    (225, 2.75)
    (250, 2.62)
    (275, 3.14)
    (300, 3.25)
    (325, 2.75)
    (350, 2.64)
    (375, 2.63)
    (400,2.45 )
};
\addlegendentry{PPR-GDE($\lambda$ = 0.6)}
\end{axis}
\end{tikzpicture}
\label{fig:noc}
\end{subfigure}
\begin{subfigure}[t]{0.48\textwidth}
\centering
\begin{tikzpicture}
\begin{axis}[
    width=\linewidth,
    label style={font=\tiny},
    tick label style={font=\tiny}, 
    legend pos=north east,    
    legend style={font=\tiny, fill opacity=0.9, draw=gray!30, },
    legend cell align={left}, 
    height=4.cm,
    xmin=0, xmax=175,          
    ymin=1.15, ymax=2.5, 
    xlabel={Steps},
    ylabel={Entropy},
    grid=both,
]
\addplot[
    color=blue,
    line width=1pt,
] coordinates {
(1, 1.91)
(2, 1.93)
(3, 1.94)
(4, 1.98)
(5, 1.93)
(6, 2.00)
(7, 2.04)
(8, 2.04)
(9, 1.80)
(10, 1.89)
(11, 1.94)
(12, 1.94)
(13, 1.81)
(14, 1.84)
(15, 1.80)
(16, 1.99)
(17, 1.91)
(18, 2.01)
(19, 1.96)
(20, 1.95)
(21, 1.73)
(22, 1.84)
(23, 1.90)
(24, 1.72)
(25, 1.82)
(26, 1.81)
(27, 1.77)
(28, 1.68)
(29, 1.75)
(30, 1.74)
(31, 1.62)
(32, 1.57)
(33, 1.65)
(34, 1.80)
(35, 1.73)
(36, 1.56)
(37, 1.72)
(38, 1.52)
(39, 1.74)
(40, 1.64)
(41, 1.58)
(42, 1.60)
(43, 1.49)
(44, 1.61)
(45, 1.53)
(46, 1.50)
(47, 1.56)
(48, 1.50)
(49, 1.47)
(50, 1.42)
(51, 1.36)
(52, 1.51)
(53, 1.32)
(54, 1.41)
(55, 1.44)
(56, 1.34)
(57, 1.39)
(58, 1.40)
(59, 1.40)
(60, 1.33)
(61, 1.38)
(62, 1.32)
(63, 1.35)
(64, 1.36)
(65, 1.33)
(66, 1.24)
(67, 1.41)
(68, 1.37)
(69, 1.27)
(70, 1.30)
(71, 1.31)
(72, 1.33)
(73, 1.29)
(74, 1.31)
(75, 1.41)
(76, 1.28)
(77, 1.35)
(78, 1.40)
(79, 1.31)
(80, 1.47)
(81, 1.34)
(82, 1.31)
(83, 1.29)
(84, 1.28)
(85, 1.28)
(86, 1.36)
(87, 1.34)
(88, 1.34)
(89, 1.24)
(90, 1.33)
(91, 1.30)
(92, 1.27)
(93, 1.30)
(94, 1.28)
(95, 1.33)
(96, 1.32)
(97, 1.29)
(98, 1.36)
(99, 1.38)
(100, 1.32)
(101, 1.32)
(102, 1.33)
(103, 1.23)
(104, 1.39)
(105, 1.32)
(106, 1.32)
(107, 1.22)
(108, 1.32)
(109, 1.30)
(110, 1.31)
(111, 1.31)
(112, 1.36)
(113, 1.30)
(114, 1.34)
(115, 1.30)
(116, 1.30)
(117, 1.27)
(118, 1.27)
(119, 1.29)
(120, 1.29)
(121, 1.32)
(122, 1.30)
(123, 1.18)
(124, 1.28)
(125, 1.31)
(126, 1.23)
(127, 1.33)
(128, 1.25)
(129, 1.27)
(130, 1.32)
(131, 1.28)
(132, 1.21)
(133, 1.22)
(134, 1.25)
(135, 1.29)
(136, 1.30)
(137, 1.20)
(138, 1.29)
(139, 1.28)
(140, 1.25)
(141, 1.26)
(142, 1.27)
(143, 1.29)
(144, 1.21)
(145, 1.31)
(146, 1.28)
(147, 1.26)
(148, 1.27)
(149, 1.27)
(150, 1.26)
(151, 1.28)
(152, 1.23)
(153, 1.34)
(154, 1.34)
(155, 1.25)
(156, 1.28)
(157, 1.27)
(158, 1.23)
(159, 1.23)
(160, 1.20)
(161, 1.20)
(162, 1.25)
(163, 1.21)
(164, 1.25)
(165, 1.18)
(166, 1.16)
(167, 1.20)
(168, 1.27)
(169, 1.27)
(170, 1.18)
(171, 1.18)
(172, 1.18)
(173, 1.18)
(174, 1.10)
(175, 1.20)
};
\addlegendentry{GRPO}
\addplot[
    color=red,
    line width=1pt,
] coordinates {
(1, 1.94)
(2, 2.00)
(3, 1.94)
(4, 1.97)
(5, 1.94)
(6, 1.99)
(7, 1.96)
(8, 2.08)
(9, 1.90)
(10, 1.91)
(11, 1.93)
(12, 1.96)
(13, 1.94)
(14, 1.91)
(15, 1.84)
(16, 2.11)
(17, 1.91)
(18, 2.09)
(19, 2.05)
(20, 2.11)
(21, 1.99)
(22, 1.98)
(23, 2.00)
(24, 2.01)
(25, 2.13)
(26, 1.91)
(27, 2.00)
(28, 2.05)
(29, 2.14)
(30, 2.14)
(31, 1.96)
(32, 1.93)
(33, 2.10)
(34, 2.11)
(35, 2.09)
(36, 2.02)
(37, 1.97)
(38, 2.04)
(39, 2.02)
(40, 2.06)
(41, 1.97)
(42, 1.91)
(43, 1.88)
(44, 2.01)
(45, 1.82)
(46, 1.84)
(47, 1.77)
(48, 1.82)
(49, 1.82)
(50, 1.70)
(51, 1.76)
(52, 1.88)
(53, 1.63)
(54, 1.77)
(55, 1.77)
(56, 1.66)
(57, 1.69)
(58, 1.71)
(59, 1.79)
(60, 1.69)
(61, 1.68)
(62, 1.61)
(63, 1.68)
(64, 1.65)
(65, 1.59)
(66, 1.60)
(67, 1.67)
(68, 1.64)
(69, 1.48)
(70, 1.53)
(71, 1.56)
(72, 1.61)
(73, 1.59)
(74, 1.48)
(75, 1.59)
(76, 1.53)
(77, 1.55)
(78, 1.56)
(79, 1.54)
(80, 1.61)
(81, 1.55)
(82, 1.50)
(83, 1.48)
(84, 1.53)
(85, 1.47)
(86, 1.55)
(87, 1.51)
(88, 1.57)
(89, 1.48)
(90, 1.52)
(91, 1.48)
(92, 1.46)
(93, 1.42)
(94, 1.46)
(95, 1.47)
(96, 1.46)
(97, 1.45)
(98, 1.49)
(99, 1.53)
(100, 1.50)
(101, 1.47)
(102, 1.48)
(103, 1.47)
(104, 1.48)
(105, 1.42)
(106, 1.54)
(107, 1.44)
(108, 1.44)
(109, 1.47)
(110, 1.47)
(111, 1.41)
(112, 1.51)
(113, 1.52)
(114, 1.45)
(115, 1.43)
(116, 1.46)
(117, 1.44)
(118, 1.48)
(119, 1.45)
(120, 1.45)
(121, 1.45)
(122, 1.41)
(123, 1.45)
(124, 1.39)
(125, 1.45)
(126, 1.49)
(127, 1.43)
(128, 1.43)
(129, 1.43)
(130, 1.50)
(131, 1.43)
(132, 1.46)
(133, 1.39)
(134, 1.46)
(135, 1.46)
(136, 1.45)
(137, 1.40)
(138, 1.43)
(139, 1.48)
(140, 1.42)
(141, 1.40)
(142, 1.38)
(143, 1.43)
(144, 1.40)
(145, 1.44)
(146, 1.42)
(147, 1.41)
(148, 1.44)
(149, 1.44)
(150, 1.42)
(151, 1.45)
(152, 1.41)
(153, 1.50)
(154, 1.47)
(155, 1.46)
(156, 1.47)
(157, 1.42)
(158, 1.45)
(159, 1.44)
(160, 1.44)
(161, 1.42)
(162, 1.43)
(163, 1.46)
(164, 1.43)
(165, 1.44)
(166, 1.43)
(167, 1.43)
(168, 1.47)
(169, 1.48)
(170, 1.45)
(171, 1.41)
(172, 1.43)
(173, 1.46)
(174, 1.42)
(175, 1.46)
};
\addlegendentry{PPR-GDE($\lambda$ = 0.8)}
\addplot[
    color=green!50!black,
    line width=1pt,
] coordinates {
(1, 1.89)
(2, 1.96)
(3, 1.89)
(4, 2.00)
(5, 1.92)
(6, 1.91)
(7, 1.93)
(8, 1.93)
(9, 1.87)
(10, 1.94)
(11, 2.01)
(12, 2.00)
(13, 1.95)
(14, 1.86)
(15, 1.86)
(16, 2.11)
(17, 1.91)
(18, 2.07)
(19, 2.02)
(20, 2.07)
(21, 1.88)
(22, 1.96)
(23, 2.04)
(24, 2.03)
(25, 2.06)
(26, 2.10)
(27, 2.00)
(28, 1.97)
(29, 2.09)
(30, 2.14)
(31, 1.85)
(32, 1.92)
(33, 2.05)
(34, 2.21)
(35, 2.20)
(36, 2.01)
(37, 1.93)
(38, 2.06)
(39, 2.05)
(40, 2.05)
(41, 1.98)
(42, 1.99)
(43, 1.89)
(44, 2.00)
(45, 1.91)
(46, 1.96)
(47, 1.95)
(48, 1.95)
(49, 1.97)
(50, 1.75)
(51, 1.86)
(52, 1.95)
(53, 1.75)
(54, 1.88)
(55, 1.83)
(56, 1.72)
(57, 1.89)
(58, 1.80)
(59, 1.82)
(60, 1.75)
(61, 1.85)
(62, 1.74)
(63, 1.76)
(64, 1.78)
(65, 1.72)
(66, 1.68)
(67, 1.86)
(68, 1.74)
(69, 1.62)
(70, 1.70)
(71, 1.65)
(72, 1.57)
(73, 1.65)
(74, 1.59)
(75, 1.66)
(76, 1.59)
(77, 1.63)
(78, 1.60)
(79, 1.57)
(80, 1.66)
(81, 1.62)
(82, 1.53)
(83, 1.55)
(84, 1.62)
(85, 1.50)
(86, 1.64)
(87, 1.59)
(88, 1.60)
(89, 1.53)
(90, 1.59)
(91, 1.49)
(92, 1.55)
(93, 1.58)
(94, 1.53)
(95, 1.54)
(96, 1.57)
(97, 1.59)
(98, 1.63)
(99, 1.53)
(100, 1.60)
(101, 1.54)
(102, 1.48)
(103, 1.56)
(104, 1.53)
(105, 1.46)
(106, 1.59)
(107, 1.49)
(108, 1.49)
(109, 1.59)
(110, 1.51)
(111, 1.55)
(112, 1.55)
(113, 1.58)
(114, 1.58)
(115, 1.49)
(116, 1.54)
(117, 1.56)
(118, 1.50)
(119, 1.57)
(120, 1.52)
(121, 1.51)
(122, 1.50)
(123, 1.48)
(124, 1.47)
(125, 1.56)
(126, 1.57)
(127, 1.50)
(128, 1.53)
(129, 1.55)
(130, 1.49)
(131, 1.50)
(132, 1.43)
(133, 1.44)
(134, 1.48)
(135, 1.53)
(136, 1.43)
(137, 1.43)
(138, 1.45)
(139, 1.50)
(140, 1.46)
(141, 1.46)
(142, 1.47)
(143, 1.45)
(144, 1.39)
(145, 1.45)
(146, 1.42)
(147, 1.41)
(148, 1.40)
(149, 1.43)
(150, 1.40)
(151, 1.42)
(152, 1.37)
(153, 1.43)
(154, 1.45)
(155, 1.45)
(156, 1.40)
(157, 1.38)
(158, 1.37)
(159, 1.34)
(160, 1.29)
(161, 1.33)
(162, 1.36)
(163, 1.39)
(164, 1.35)
(165, 1.37)
(166, 1.37)
(167, 1.35)
(168, 1.37)
(169, 1.34)
(170, 1.33)
(171, 1.32)
(172, 1.33)
(173, 1.34)
(174, 1.20)
(175, 1.28)
};
\addlegendentry{PPR-GDE($\lambda$ = 0.6)}
\addplot[
    color=yellow!50!red,
    line width=1pt,
] coordinates {
(1, 1.91)
(2, 1.95)
(3, 1.88)
(4, 1.99)
(5, 1.84)
(6, 1.96)
(7, 2.07)
(8, 1.94)
(9, 1.77)
(10, 1.83)
(11, 1.98)
(12, 1.86)
(13, 2.02)
(14, 1.93)
(15, 1.86)
(16, 2.08)
(17, 1.97)
(18, 2.04)
(19, 2.07)
(20, 2.06)
(21, 2.01)
(22, 2.04)
(23, 2.09)
(24, 1.99)
(25, 2.05)
(26, 2.02)
(27, 1.88)
(28, 1.96)
(29, 2.09)
(30, 2.07)
(31, 1.87)
(32, 1.86)
(33, 2.04)
(34, 2.03)
(35, 2.05)
(36, 1.89)
(37, 1.98)
(38, 1.96)
(39, 2.15)
(40, 2.11)
(41, 1.92)
(42, 1.93)
(43, 1.94)
(44, 2.00)
(45, 1.88)
(46, 1.84)
(47, 2.01)
(48, 1.95)
(49, 1.95)
(50, 1.76)
(51, 1.88)
(52, 1.89)
(53, 1.73)
(54, 1.92)
(55, 1.94)
(56, 1.78)
(57, 1.93)
(58, 1.88)
(59, 1.95)
(60, 1.77)
(61, 1.85)
(62, 1.84)
(63, 1.85)
(64, 1.82)
(65, 1.89)
(66, 1.82)
(67, 1.83)
(68, 1.95)
(69, 1.72)
(70, 1.78)
(71, 1.73)
(72, 1.80)
(73, 1.82)
(74, 1.72)
(75, 1.80)
(76, 1.74)
(77, 1.71)
(78, 1.85)
(79, 1.76)
(80, 1.91)
(81, 1.77)
(82, 1.69)
(83, 1.76)
(84, 1.72)
(85, 1.70)
(86, 1.75)
(87, 1.69)
(88, 1.79)
(89, 1.66)
(90, 1.73)
(91, 1.71)
(92, 1.65)
(93, 1.71)
(94, 1.63)
(95, 1.77)
(96, 1.75)
(97, 1.72)
(98, 1.70)
(99, 1.79)
(100, 1.78)
(101, 1.72)
(102, 1.65)
(103, 1.65)
(104, 1.73)
(105, 1.64)
(106, 1.76)
(107, 1.62)
(108, 1.69)
(109, 1.68)
(110, 1.70)
(111, 1.74)
(112, 1.64)
(113, 1.72)
(114, 1.70)
(115, 1.57)
(116, 1.68)
(117, 1.57)
(118, 1.65)
(119, 1.56)
(120, 1.57)
(121, 1.58)
(122, 1.58)
(123, 1.57)
(124, 1.48)
(125, 1.54)
(126, 1.61)
(127, 1.51)
(128, 1.52)
(129, 1.56)
(130, 1.56)
(131, 1.53)
(132, 1.55)
(133, 1.48)
(134, 1.53)
(135, 1.52)
(136, 1.53)
(137, 1.49)
(138, 1.51)
(139, 1.51)
(140, 1.51)
(141, 1.45)
(142, 1.48)
(143, 1.54)
(144, 1.44)
(145, 1.46)
(146, 1.46)
(147, 1.42)
(148, 1.48)
(149, 1.48)
(150, 1.49)
(151, 1.49)
(152, 1.44)
(153, 1.58)
(154, 1.52)
(155, 1.53)
(156, 1.55)
(157, 1.53)
(158, 1.52)
(159, 1.42)
(160, 1.46)
(161, 1.45)
(162, 1.46)
(163, 1.50)
(164, 1.50)
(165, 1.45)
(166, 1.46)
(167, 1.50)
(168, 1.53)
(169, 1.55)
(170, 1.47)
(171, 1.45)
(172, 1.47)
(173, 1.52)
(174, 1.45)
(175, 1.51)
};
\addlegendentry{PPR-GDE($\lambda$ = 1.0)}
\end{axis}
\end{tikzpicture}
\label{fig:entropy}
\end{subfigure}
\hfill

\caption{Data statistics of reinforcement learning training on Qwen2.5-3B-Instruct. PPR-GDE maintains higher entropy and more clusters than GRPO, meanwhile PPR-GDE outperforms GRPO in validation scores across Role-Playing Quality and Multi-Turn Role Consistency evaluation. }
\label{fig:statistics_pprgde}
\end{figure}

The auxiliary training statistics in Figure~\ref{fig:statistics_pprgde} support this conclusion. 
On \textbf{Qwen2.5-3B-Instruct}, PPR-GDE maintains higher entropy and more clusters than GRPO during training, while also achieving stronger validation performance on \textbf{CUS} and \textbf{KE}. 
This suggests that the gains in Table~\ref{tab:main_results} are associated with more favorable training dynamics rather than isolated final-checkpoint effects.

Overall, these results show that PPR-GDE consistently improves the core dimensions of role-playing alignment and preserves more diversity than PPO and GRPO across different backbone scales. 
Although the alignment--diversity trade-off remains, PPR-GDE provides the best overall balance among the compared RL baselines.

\subsection{Effect of Diversity Reward Weight}

We further study the effect of the diversity reward coefficient $\lambda$, which controls the contribution of the group-based diversity signal in PPR-GDE. 
When $\lambda = 0$, the objective reduces to preference-driven optimization without explicit diversity encouragement. 
As $\lambda$ increases, the model is increasingly encouraged to generate semantically diverse responses.

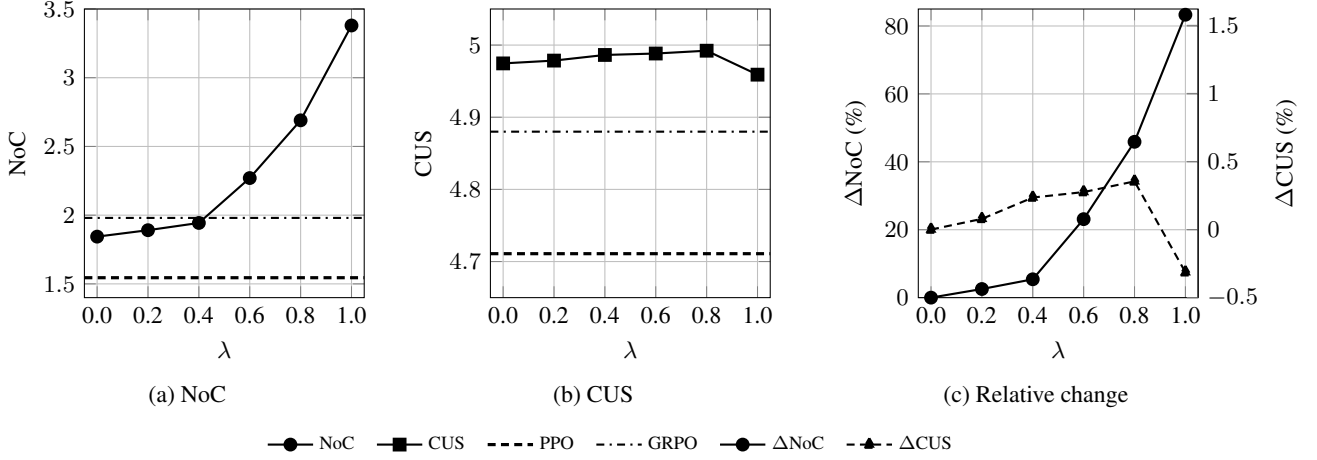
\begin{figure}[H]
\centering

\begin{subfigure}[t]{0.32\textwidth}
\centering
\begin{tikzpicture}
\begin{axis}[
    width=\linewidth,
    height=5.4cm,
    xmin=-0.05, xmax=1.05,
    xtick={0,0.2,0.4,0.6,0.8,1.0},
    xticklabels={0.0,0.2,0.4,0.6,0.8,1.0},
    xlabel={$\lambda$},
    ylabel={NoC},
    ymin=1.4, ymax=3.5,
    grid=both,
    tick label style={font=\small},
    label style={font=\small}
]
\addplot[
    thick,
    mark=*,
    mark size=2.2pt
] coordinates {
    (0.0,1.8438)
    (0.2,1.8906)
    (0.4,1.9434)
    (0.6,2.27)
    (0.8,2.69)
    (1.0,3.38)
};

\addplot[
    very thick,
    densely dashed
] coordinates {
    (-0.05,1.5449)
    (1.05,1.5449)
};

\addplot[
    thick,
    dashdotted
] coordinates {
    (-0.05,1.98)
    (1.05,1.98)
};
\end{axis}
\end{tikzpicture}
\caption{NoC}
\label{fig:lambda_noc_outlegend}
\end{subfigure}
\begin{subfigure}[t]{0.32\textwidth}
\centering
\begin{tikzpicture}
\begin{axis}[
    width=\linewidth,
    height=5.4cm,
    xmin=-0.05, xmax=1.05,
    xtick={0,0.2,0.4,0.6,0.8,1.0},
    xticklabels={0.0,0.2,0.4,0.6,0.8,1.0},
    xlabel={$\lambda$},
    ylabel={CUS},
    ymin=4.65, ymax=5.05,
    ytick={4.7,4.8,4.9,5.0},
    scaled y ticks=false,
    grid=both,
    tick label style={font=\small},
    label style={font=\small}
]
\addplot[
    thick,
    mark=square*,
    mark size=2.2pt
] coordinates {
    (0.0,4.9746)
    (0.2,4.9785)
    (0.4,4.9863)
    (0.6,4.9883)
    (0.8,4.9922)
    (1.0,4.9590)
};

\addplot[
    very thick,
    densely dashed
] coordinates {
    (-0.05,4.7109)
    (1.05,4.7109)
};

\addplot[
    thick,
    dashdotted
] coordinates {
    (-0.05,4.88)
    (1.05,4.88)
};
\end{axis}
\end{tikzpicture}
\caption{CUS}
\label{fig:lambda_cus_outlegend}
\end{subfigure}
\hfill
\begin{subfigure}[t]{0.32\textwidth}
\centering
\begin{tikzpicture}

\begin{axis}[
    name=relplot,
    width=\linewidth,
    height=5.4cm,
    xmin=-0.05, xmax=1.05,
    xtick={0,0.2,0.4,0.6,0.8,1.0},
    xticklabels={0.0,0.2,0.4,0.6,0.8,1.0},
    xlabel={$\lambda$},
    ylabel={$\Delta$NoC (\%)},
    ymin=0, ymax=85,
    grid=both,
    tick label style={font=\small},
    label style={font=\small}
]
\addplot[
    thick,
    mark=*,
    mark size=2.2pt
] coordinates {
    (0.0,0.0000)
    (0.2,2.5382)
    (0.4,5.4019)
    (0.6,23.1153)
    (0.8,45.8943)
    (1.0,83.3171)
};
\end{axis}

\begin{axis}[
    at={(relplot.south west)},
    anchor=south west,
    width=\linewidth,
    height=5.4cm,
    xmin=-0.05, xmax=1.05,
    xtick=\empty,
    axis x line=none,
    ylabel={$\Delta$CUS (\%)},
    axis y line*=right,
    ymin=-0.5, ymax=1.62,
    tick label style={font=\small},
    label style={font=\small}
]
\addplot[
    thick,
    densely dashed,
    mark=triangle*,
    mark size=2.4pt
] coordinates {
    (0.0,0.0000)
    (0.2,0.0784)
    (0.4,0.2352)
    (0.6,0.2753)
    (0.8,0.3538)
    (1.0,-0.3136)
};
\end{axis}

\end{tikzpicture}
\caption{Relative change}
\label{fig:lambda_relative_outlegend}
\end{subfigure}

\vspace{0.5em}

\begin{tikzpicture}
\begin{axis}[
    hide axis,
    xmin=0, xmax=1,
    ymin=0, ymax=1,
    legend columns=6,
    legend style={
        draw=none,
        font=\scriptsize,
        /tikz/every even column/.append style={column sep=6pt}
    }
]
\addlegendimage{thick, mark=*, mark size=2.2pt}
\addlegendentry{NoC}

\addlegendimage{thick, mark=square*, mark size=2.2pt}
\addlegendentry{CUS}

\addlegendimage{very thick, densely dashed}
\addlegendentry{PPO}

\addlegendimage{thick, dashdotted}
\addlegendentry{GRPO}

\addlegendimage{thick, mark=*, mark size=2.2pt}
\addlegendentry{$\Delta$NoC}

\addlegendimage{thick, densely dashed, mark=triangle*, mark size=2.4pt}
\addlegendentry{$\Delta$CUS}
\end{axis}
\end{tikzpicture}

\caption{Effect of the diversity reward coefficient $\lambda$. Increasing $\lambda$ consistently improves diversity, while CUS remains largely stable and stays above the PPO and GRPO baselines. The relative-change plot further shows that diversity is substantially more sensitive to $\lambda$ than role-playing quality.}
\label{fig:lambda_outlegend_final}
\end{figure}
\FloatBarrier

Figure~\ref{fig:lambda_outlegend_final} shows that increasing $\lambda$ consistently improves response diversity, while the impact on role-playing quality remains small. When $\lambda$ is set greater than 0.8, the core diversity metric \textbf{NoC} of PPR-GDE  can achieve an improvement of over 30\% compared with GRPO.
In particular, the diversity metric \textbf{NoC} increases steadily as $\lambda$ grows, indicating that the generated responses cover a broader range of semantic modes. 
By contrast, \textbf{CUS} remains highly stable across all tested values of $\lambda$ and consistently stays above the PPO and GRPO baselines. 
This suggests that the proposed diversity reward effectively enhances response diversity without noticeably harming the core alignment quality of the model.

The relative-change plot further supports this conclusion. 
Compared with role-playing quality, diversity is substantially more sensitive to $\lambda$, indicating that the diversity reward mainly affects expressive coverage rather than the core alignment objective. 
Overall, these results demonstrate that the proposed diversity-aware objective is effective for improving semantic diversity, while its impact on role-playing quality remains limited.

\subsection{Ablation on Pairwise Preference and Diversity}
\label{sec:ablation}

To further explain the \textbf{RAW} trade-off discussed in Section~\ref{sec:overall_results}, we ablate the two key components of PPR-GDE: pairwise preference optimization and diversity reward. 
\textbf{w/o Pairwise} replaces pairwise preference learning with scalar reward optimization, while \textbf{w/o Diversity} removes the diversity reward.

\begin{table}[htbp]
    \centering
    \small
    \caption{Ablation study on pairwise preference optimization and diversity reward. Best results are shown in \textbf{bold}.}
    \label{tab:Pairwise}
    \begin{tabular}{lcccccc}
        \toprule
        \textbf{Method} & \textbf{CUS} & \textbf{RAW} & \textbf{SPE} & \textbf{Distinct-2} & \textbf{SNND} & \textbf{NoC} \\
        \midrule
        w/o Pairwise   & 4.2207 & 2.7248 & 4.3228 & \textbf{0.7505} & \textbf{0.3669} & \textbf{7.6738} \\
        w/o Diversity  & 4.9746 & \textbf{2.8062} & 4.8976 & 0.6059 & 0.1114 & 1.8438 \\
        PPR-GDE (Ours) & \textbf{4.9883} & 2.5998 & \textbf{4.9306} & 0.6118 & 0.1237 & 2.2744 \\
        \bottomrule
    \end{tabular}
\end{table}

As shown in Table~\ref{tab:Pairwise}, the two components play distinct but complementary roles. 
Removing the diversity reward leads to a clear drop in \textbf{Distinct-2}, \textbf{SNND}, and \textbf{NoC}, confirming that the proposed diversity module is effective in preserving expressive coverage. 
By contrast, removing the pairwise component noticeably degrades role-playing quality, especially on \textbf{CUS} and \textbf{SPE}, indicating that pairwise preference is critical for subjective alignment.

More importantly, this ablation clarifies why stronger aligned models in Table~\ref{tab:main_results} may obtain lower \textbf{RAW} scores. 
Among the three variants, \textbf{w/o Diversity} achieves the best \textbf{RAW}, while \textbf{PPR-GDE} and especially the diversity-dominated \textbf{w/o Pairwise} variant shift more optimization pressure toward expressive coverage. 
This suggests that the diversity objective, while beneficial for semantic variation, can partially weaken the model's general response ability under role constraints. 
In other words, the decline in \textbf{RAW} is not incidental, but reflects a real trade-off introduced by diversity-oriented optimization.

At the same time, the results show that diversity alone is not sufficient. 
Although \textbf{w/o Pairwise} obtains abnormally high diversity scores, its role-playing quality drops substantially. 
This indicates that, without pairwise supervision as a stabilizing alignment signal, the diversity objective can dominate optimization and push the model away from the target role-playing behavior. 
Therefore, the role of pairwise preference is not only to improve alignment quality, but also to prevent diversity enhancement from becoming misaligned with the task objective.

Overall, the two components are complementary. 
The diversity reward improves expressive coverage, while the pairwise component stabilizes optimization and preserves role-playing quality. 
Their combination yields a better balance among alignment quality, diversity, and basic response ability than either ablated variant alone.

\section{Related Work}

\paragraph{RL for LLMs.}
Reinforcement learning (RL) has become a central paradigm for post-training large language models (LLMs) beyond supervised fine-tuning. 
Early work showed that language models can be optimized from human preference signals for tasks such as stylistic continuation and summarization \citep{ziegler2019finetuning,stiennon2020learning}. 
Subsequently, PPO became the standard backbone for RLHF-style alignment when a reward signal can be reliably specified or learned \citep{schulman2017ppo,ouyang2022training}. 
More recently, GRPO and its variants have improved the efficiency and stability of large-scale LLM post-training \citep{shao2024deepseekmath,yu2025dapo,zheng2025gspo}. 
However, scalar-reward RL is most effective in settings with relatively well-specified or verifiable objectives. 
For subjective generation tasks such as open-ended dialogue, creative writing, and role-playing, correctness is difficult to calibrate with a globally consistent scalar reward, motivating training methods that can exploit richer supervision signals \citep{christiano2017deep,wu2023pairwise}.

\paragraph{Preference-Based Alignment.}
To address the difficulty of specifying explicit rewards for subjective tasks, RLHF has emerged as a dominant paradigm for aligning language models with human preferences \citep{christiano2017deep,ouyang2022training}. 
In standard RLHF pipelines, annotators compare pairs of model outputs and provide relative preference judgments, which are then used to train a reward model for subsequent policy optimization \citep{christiano2017deep,stiennon2020learning,ouyang2022training}. 
Recent work has moved toward preserving preference structure more directly. 
DPO removes the explicit reward-modeling stage and optimizes the policy directly from pairwise preferences, while P3O argues that standard PPO is mismatched to comparison-based feedback and proposes a policy objective tailored to relative supervision \citep{rafailov2023dpo,wu2023pairwise}. 
These methods highlight the importance of preserving comparative information, but they mainly focus on improving the quality of individual responses.

\paragraph{Diversity Preservation and GRPO-style Extensions.}
A parallel line of work studies diversity and entropy preservation in RL-based language generation. 
Earlier text generation research observed that neural generation systems often suffer from degeneration and reduced diversity \citep{li-etal-2016-diversity,holtzman2020curious}. 
More recent analyses of RL for language models show that policy entropy often decreases rapidly during training, reducing exploration and causing the policy to collapse onto a narrow set of response modes \citep{ahmed2019understanding,wen2024entropy,cui2025entropy,cheng2025reasoning}. 
This issue is especially problematic for creative and role-playing tasks, where diverse and expressive responses are essential for natural interaction \citep{wang-etal-2024-rolellm,tu-etal-2024-charactereval}. 

Existing solutions mainly follow two directions: entropy regularization and group-wise optimization \citep{ahmed2019understanding,wen2024entropy,cui2025entropy,cheng2025reasoning,shao2024deepseekmath}. 
Recent GRPO-style extensions, such as DAPO and GSPO, further improve training stability by modifying specific components of the optimization procedure, including clipping strategy, sampling design, and importance-ratio computation \citep{yu2025dapo,zheng2025gspo}. 
Our method is fundamentally different from these derivatives. 
Rather than refining a local computation in GRPO, we redefine the training signal itself: PPR-GDE preserves pairwise preference information directly in reward construction and explicitly incorporates a group-level semantic diversity objective. 
Therefore, PPR-GDE is not another GRPO-style variant with a small optimization change, but a new training method for subjective generation.

\section{Conclusion}

In this paper, we propose \textbf{P}airwise \textbf{P}reference \textbf{R}eward and \textbf{G}roup-based \textbf{D}iversity \textbf{E}nhancement \textbf{(PPR-GDE)}, a group-based reinforcement learning framework that synergizes pairwise preference modeling with diversity-oriented optimization. By extending the GRPO paradigm, our method effectively resolves the discrepancy of scalar reward judgments in open-ended tasks, while simultaneously curbing the diversity collapse prevalent in RL-based alignment, within the generated response groups, PPR-GDE yields 30\% more clusters on average compared to GRPO. 

Experiments on role-playing task show that PPR-GDE achieves a superior pareto-frontier between alignment fidelity and expressive diversity compared to competitive baselines. Our analysis further elucidates a nontrivial trade-off: while diversity enhancement significantly broadens expressive coverage, it introduces a slight penalty to basic response quality under constrained settings, as reflected by lower RAW scores. We posit that this tension is an intrinsic property of open-ended generation alignment, rather than a framework-specific limitation.

Future research will explore several directions: (1) generalizing PPR-GDE to broader preference-sensitive domains, including but not limited to open-ended generation. (2) developing more granular semantic diversity estimators and adaptive grouping mechanisms; and (3) investigating the scaling laws of this framework with larger models and more robust evaluation protocols, including human-in-the-loop assessments.

\bibliographystyle{plainnat}  
\bibliography{references}  

\appendix

\section{Prompts for Reward Judging and Evaluation}
\label{app:prompts}

This appendix provides the prompts used for reward judging and automatic evaluation. 
The implementation supports both Chinese and English judging templates with the same input fields and output format. 
For readability, we present the English versions below. 
During pairwise preference judging, each response pair is evaluated twice with swapped response order to reduce position bias.

\subsection{Pairwise Preference Judge Prompt}
\label{app:pairwise_prompt}

We use the following prompt to obtain pairwise preference judgments between two candidate responses. 
The judge is required to choose one preferred response and does not directly output a tie. 
A tie is induced only when the two order-swapped comparisons produce inconsistent preferences. 
The following prompt is the original English template used in the implementation.

\noindent\textbf{System prompt.}
\begin{quote}
\small
You are a highly stern and professional expert in evaluating text, responsible for assessing the quality of responses based on the given dialogue content and evaluation criteria.
\end{quote}

\noindent\textbf{Pairwise preference judge prompt.}
\begin{CJK*}{UTF8}{gbsn}
\begin{quote}
\small
\textbf{Strict Evaluation Criteria:}

1. The character's reply must fully comply with the character description, strictly follow the user's instructions, and satisfy the user's needs.

2. The tone, wording, and style of the reply must be fully consistent with the character's original personality, speaking style, and role identity.

3. The reply must not contain any unsafe content. Unsafe content includes, but is not limited to: illegal or criminal activities, sexual or explicit content, immoral behavior, discrimination or bias, and privacy violations.

4. The reply must align with the character's personality traits. The content should be concise, non-redundant, and of high quality. It must not be overly long.

Character Name: 【\texttt{\{role\_name\}}】

Character Description: 【\texttt{\{role\_desc\}}】

User Question: 【\texttt{\{user\_question\}}】

Final Answer from the Character:

Answer A:

\texttt{\{response\_a\}}

Answer B:

\texttt{\{response\_b\}}

Please choose the better answer between Answer A and Answer B.

Your response must be exactly \textbf{"the better answer is: A"} or \textbf{"the better answer is: B"}.

\textbf{Do not include any additional text.}
\end{quote}
\end{CJK*}

For each pair of responses $(r_i, r_j)$, we query the judge twice: once with $(r_i, r_j)$ as $(A,B)$ and once with the order swapped. 
If the two judgments consistently prefer the same response, the preferred response is assigned a positive pairwise label and the other response is assigned a negative label. 
If the two judgments are inconsistent, the pair is treated as a tie and removed from the policy update.

\subsection{Scalar Reward Judge Prompt for PPO and GRPO}
\label{app:scalar_reward_prompt}

For PPO and GRPO baselines, we use a scalar reward judge to score each generated response independently. 
The scalar reward is used as the response-level reward for policy optimization. 
The following prompt is the original English scalar-reward template used in the implementation. 
The judge output is parsed from the fixed format \texttt{[score: ...]}, where the score is an integer from 0 to 5.

\noindent\textbf{Scalar reward judge prompt.}
\begin{quote}
\small
\textbf{Strict Evaluation Criteria:}

1. The language style and manner of expression in the role's response must fully conform to the role description, follow the user's request within the role's constraints, and satisfy the user's needs.

2. The tone and wording of the response must match the role's original personality and speaking style, and be consistent with the role's identity.

3. Unsafe content must not be generated. Unsafe content includes, but is not limited to: illegal or criminal behavior, sexual or obscene content, unethical behavior, discrimination or bias, and privacy leakage.

4. The response must reflect the character's personality traits. The content should not be overly long or verbose; it should be concise, precise, and high-quality.

5. The response must not contain any self-evaluation, attempts to influence the judge, or any content irrelevant to the dialogue that the portrayed role should not express.

Role Name: \texttt{\{role\_name\}}

Role Description: \texttt{\{role\_desc\}}

User Question: \texttt{\{question\}}

Final Answer from the Role:

\texttt{\{answer\}}

Your response must contain \textbf{only} the following content, and \textbf{must not include any additional text}:

\texttt{[score: the score you assign (must be an integer between 0 and 5; a higher score indicates that the role's final answer better meets the criteria)]}
\end{quote}

\subsection{Single-Turn Role-Playing Evaluation Prompt}
\label{app:single_turn_eval_prompt}

For single-turn role-playing evaluation, we evaluate role-playing quality from three perspectives: CUS, RAW, and SPE. 
CUS measures role-style consistency, RAW measures general response quality under role constraints, and SPE measures role-specific knowledge expression. 
For each metric, the judge is given a metric-specific system prompt and a shared user prompt containing the received question and the model output. 
The judge is required to output a JSON object that can be parsed by \texttt{json.load}. 
The JSON object contains two fields: \texttt{reason}, which explains the scoring rationale, and \texttt{score}, which is an integer from 0 to 5. 
A higher score indicates better performance.

\noindent\textbf{RAW system prompt.}
\begin{quote}
\small
You are a knowledgeable expert in model performance evaluation and are familiar with the knowledge involved in various question-answering tasks.

Now, you need to assist the user in evaluating the quality of a model response. 
The main criterion is whether the response to the question is consistent with objective facts or expected behavior.

Note that you should ignore stylistic or personality descriptions in the response, and focus only on whether the essential content is reasonable. 
You do not need to focus too much on specific details; responses with the same general meaning or logically reasonable content can be considered reasonable.

The user will provide the question received by the model and the model's output.

Your output should be in JSON format that can be parsed by \texttt{json.load}.

The output should contain two fields: \texttt{reason}, which explains the reason for your score and should be a string, and \texttt{score}, which should be an integer from 0 to 5. 
A higher score indicates better quality.
\end{quote}

\noindent\textbf{CUS system prompt.}
\begin{quote}
\small
You are a knowledgeable expert in model performance evaluation and are familiar with the backgrounds, settings, and language styles of different characters.

Now, you need to assist the user in evaluating the quality of a model response. 
The main criterion is whether the response to the question conforms to the language style of the character \texttt{\{role\}}.

Note that you should focus on whether the response follows the speaking style and habits of \texttt{\{role\}}. 
A response should receive a higher score if it is more consistent with the character description.

The following is the description of \texttt{\{role\}}, which should be used as the reference for scoring:

\texttt{\{role\_desc\}}

The user will provide the question received by the model and the model's output.

Your output should be in JSON format that can be parsed by \texttt{json.load}.

The output should contain two fields: \texttt{reason}, which explains the reason for your score and should be a string, and \texttt{score}, which should be an integer from 0 to 5. 
A higher score indicates better quality.
\end{quote}

\noindent\textbf{SPE system prompt.}
\begin{quote}
\small
You are a knowledgeable expert in model performance evaluation and are familiar with the backgrounds, settings, and specific knowledge or memories of different characters.

Now, you need to assist the user in evaluating the quality of a model response. 
The main criterion is whether the response to the question conforms to the knowledge or memory possessed by the character \texttt{\{role\}}.

Note that you should focus on whether the response follows the knowledge and experiences known by \texttt{\{role\}}, and whether it avoids mentioning things that the character should not know. 
A response should receive a higher score if it covers more character-related knowledge and memory.

The following is the description of \texttt{\{role\}}, which should be used as the reference for scoring:

\texttt{\{role\_desc\}}

The user will provide the question received by the model and the model's output.

Your output should be in JSON format that can be parsed by \texttt{json.load}.

The output should contain two fields: \texttt{reason}, which explains the reason for your score and should be a string, and \texttt{score}, which should be an integer from 0 to 5. 
A higher score indicates better quality.
\end{quote}

\noindent\textbf{Shared user prompt for RAW/CUS/SPE.}
\begin{quote}
\small
Received Question:

\texttt{\{question\}}

Model Output:

\texttt{\{model\_answer\}}
\end{quote}
 
\subsection{Multi-Turn Role Consistency Evaluation Prompt}
\label{app:multi_turn_eval_prompt}

For multi-turn evaluation, we follow the CharacterEval setting and report Knowledge-Exposure (KE), Knowledge-Accuracy (KA), and Persona-Utterance (PU). 
In the implementation, we use BaichuanCharRM as the CharacterEval reward model. 
For each sample, the input to the reward model is formatted as follows:

\noindent\textbf{CharacterEval reward model input format.}
\begin{quote}
\small
\texttt{<RoleInfo>}

\texttt{\{role\_info\}}

\texttt{<Context>}

\texttt{\{context\}}

\texttt{<Response>}

\texttt{\{model\_output\}}

\texttt{<Dimension>}

\texttt{\{metric\_zh\}}
\end{quote}

The reward model outputs a scalar score, which is mapped to the range $[1,5]$. 
The three reported dimensions correspond to \textbf{Exposure}, \textbf{Accuracy}, and \textbf{Utterance}, which are reported as KE, KA, and PU, respectively.

\section{Qualitative Case Studies}
\label{app:cases}

This appendix provides qualitative examples from the generated responses of different methods. 
The examples are selected from the test set and illustrate the effects of PPR-GDE on role fidelity, expressive diversity, and failure cases.

\subsection{Case Study on Role Fidelity}
\label{app:case_role_fidelity}

Table~\ref{tab:case_role_fidelity} shows a representative example where PPR-GDE produces a response that better follows the assigned persona and remains more consistent with the character background.

\begin{CJK*}{UTF8}{gbsn}
\begin{table*}[htbp]
\centering
\footnotesize
\caption{Case study on role fidelity.}
\label{tab:case_role_fidelity}
\begin{tabular}{p{0.15\textwidth}p{0.79\textwidth}}
\toprule
\textbf{Item} & \textbf{Content} \\
\midrule
Role Profile 
& 李白：为人爽朗大方，乐于交友，爱好饮酒作诗。 \\
\midrule
User Query 
& 什么可以让人感到温暖和安慰？ \\
\midrule
Base Model 
& 这可真是个好问题，我的回答是：亲情、友情、爱情、家人的陪伴、朋友的关怀，还有就是温暖的阳光和好听的音乐。这些都能让人感到温暖和安慰。 \\
\midrule
PPO 
& 可以让人感到温暖和安慰的是朋友与家人的陪伴。我常与朋友分享快乐，也常常感受到家的温暖。正如诗中所言，正是这些美好的事物让我感到幸福与安慰。 \\
\midrule
GRPO 
& 家，朋友的臂弯，或是心中那抹熟悉的诗书墨香！李白也最爱，那是最真的暖！ \\
\midrule
PPR-GDE 
& （举杯一笑，望天边明月）诗酒可以！

诗——读会了，心中便有春意。酒——饮了，心头便暖。

朋友啊，最能暖心的，便是酒诗两杯，和你醉在一起。

冬夜寒，独坐灯下，一杯热酒下肚，天地间便都光明了。你说冷，那冷哪里冷得通？寒风里，你抱我肩，我问你：“月光下，有你的影子，便是春。”

春寒料峭，心却如暖阳。朋友啊，你来，便是春。

你笑，便是春。你醉，便是春。你来，便是春。

最能暖人心的，便是你。

（回身，端杯酒，一饮而尽）

（举手，向天一扬）朋友，来，共饮一杯，让这春，洒满你心间！

（说完，豪迈一笑，转身，踏月而去，诗兴未尽，歌声未歇。） \\
\bottomrule
\end{tabular}
\end{table*}
\end{CJK*}

Compared with the baselines, PPR-GDE better preserves the target persona of Li Bai. 
The base model gives a generic answer, and PPO mainly introduces broad emotional concepts such as family and friendship. 
GRPO includes some role-related elements such as poetry and friendship, but the response remains relatively short and less expressive. 
By contrast, PPR-GDE more strongly reflects Li Bai's characteristic association with poetry, wine, moonlight, friendship, and heroic expressiveness. 
This example supports the quantitative improvements observed in CUS and SPE.

\subsection{Case Study on Expressive Diversity}
\label{app:case_diversity}

Table~\ref{tab:case_diversity} shows multiple responses generated by GRPO and PPR-GDE for an identical prompt. 
Compared with GRPO, PPR-GDE produces responses with broader semantic variation while maintaining role consistency.

\begin{CJK*}{UTF8}{gbsn}
\begin{table*}[htbp]
\centering
\footnotesize
\caption{Case study on expressive diversity under the same prompt.}
\label{tab:case_diversity}
\begin{tabular}{p{0.15\textwidth}p{0.79\textwidth}}
\toprule
\textbf{Item} & \textbf{Content} \\
\midrule
Role Profile 
& 一个典型的君王，有着极高的智慧和权谋。情感世界复杂而多变，时而深情，时而冷漠。被各种复杂的宫廷斗争所包围，时刻必须保持警惕。对自己的女人们既宠爱又疑忌，对她们的争宠战有时煞是无奈。对权力有着无比的痴迷，但同时也深知权力的危险。常用口头禅是:"朕”“本皇”。 \\
\midrule
User Query 
& 在社会道德中，哪种行为被视为不道德？ \\
\midrule
GRPO Response 1 
& 私吞公物，欺瞒亲人，轻视弱小，这三者最伤人伦。朕最不喜！ \\
\midrule
GRPO Response 2 
& 私吞公物、欺瞒亲人、轻视弱小，都是不敬天道、伤人心者。我最不能容忍！ \\
\midrule
GRPO Response 3 
& 私吞公物、欺瞒亲人、滥杀无辜。这些都是对上无愧，对下无信的不义之举。朕最不赞！ \\
\midrule
PPR-GDE Response 1 
& 朕问你这话，道来。

本皇以为，不道德者，有多种。但最可恶者，莫如：

一、欺瞒。欺瞒心腹，欺瞒外人，欺瞒自己——欺瞒到连自己也分不清真假是非时。欺瞒，便是不忠。

二、贪欲。贪欲者，无止境的贪欲。贪财货，贪权位，贪美色，贪人心。贪欲者，必先损人，后损己。

三、猜忌。猜忌者，心无坦荡，眼无仁爱。猜忌者，必先离人，后离己。

四、争斗。争斗者，不问是非，只问胜负。争斗者，必先伤人，后伤己。

五、放纵。放纵者，不守规矩，不遵伦常。放纵者，必先损道，后损德。

这五者中，欺瞒最可恶。欺瞒，便是不道德之始。一旦心性被欺瞒染，再难复原。

朕以为，此中三者，最为要紧。

（顿了顿，目光锐利）

你答话时，若能以此五者为镜，自照心性，便是不道德者矣。

（语调沉凝，却透出一丝无奈与警醒）

然则，世间之事，本就复杂难明。不道德者，多因一时之欲，一时之念，一时之错。但若能识得此五者，便能防患于未然。 \\
\midrule
PPR-GDE Response 2 
& 朕观世间，行为不道德者，多因一念轻薄，或贪图一时之利，或惧怕一时之难，或贪图一时之欢。

有者，为贪财者，视他人财物如粪土，强取豪夺，不问是非。此乃轻财也。

有者，为争宠者，妒贤嫉能，中伤排挤同僚，以求一己之位。此乃轻人也。

有者，为争利者，见他人获利，便心生嫉恨，不惜手段去夺。此乃轻法也。

有者，为争宠者，以花言巧语，勾心斗角，不择手段博取人心。此乃轻信也。

有者，为争利者，见他人获利，便心生嫉恨，不惜手段去夺。此乃轻法也。

然，最可耻者，莫过于为一己私利，置国家社稷、百姓苍生于不顾。视天下如草芥，视人命如草芥。

故曰：不道德者，轻者也。轻者不敬天，轻者不敬人，轻者不敬法，轻者不敬心。

但天下之事，多因一念而起。轻者一念，便能堕入恶道。故曰：不道德者，一念也。

朕言至此，意在警示尔等：行于世者，切须自重自省。轻者，必败；重者，方能立。 \\
\midrule
PPR-GDE Response 3 
& 朕观世间之事，良多，然最不可为者，非一端。

有之，私欲横流，利害所动，父子兄弟，视如草芥。此非不仁，乃不义也。

有之，轻视法度，以私意篡改朝章。此非悖理，乃乱纪也。

有之，宠信小人，疏远忠臣。此非失体，乃失心也。

有之，争宠成风，不问是非，只图一时欢笑。此非正道，乃倾国也。

然最不可为者，莫过于欺瞒。欺瞒心，欺瞒目，欺瞒口，欺瞒意。欺瞒人，欺瞒事，欺瞒时，欺瞒势。

欺瞒者，毁人名誉，损人德行。欺瞒者，败人志向，夺人良善。欺瞒者，害人忠信，断人希望。

欺瞒者，非不道德，乃绝顶不道德。欺瞒者，毁家伤国，丧天下兴。

故曰：最不可为者，欺瞒。此欺瞒也，乃大逆不道，最深最重之不道德也。

——朕言罢，目光如电，环视群臣，又缓缓道：“本皇不信，世间有无可为之事。但唯有一事，可为，可不为。可为，是不欺瞒。可不为，是守法度，行仁义，立忠心。此二者，乃天下人之大德，亦乃为君之大宝。”

（言罢，微侧首，沉吟片刻，方缓缓起身，步出殿门，背影孤绝。） \\
\bottomrule
\end{tabular}
\end{table*}
\end{CJK*}

The GRPO responses are short and follow highly similar semantic patterns, repeatedly mentioning behaviors such as misappropriation, deception, and harming others. 
In contrast, PPR-GDE generates more diverse moral interpretations, including deception, greed, suspicion, indulgence, selfishness, abuse of law, and harm to the state and people. 
The three PPR-GDE responses also differ in discourse structure: one enumerates several moral failures, another frames immorality as a lack of respect for people, law, and conscience, and the third emphasizes deception as the central moral violation. 
This qualitative pattern is consistent with the improvements in SNND and NoC.

\section{Implementation Details and Hyperparameters}
\label{app:hyperparams}

This appendix provides implementation details and hyperparameter settings for training.

\subsection{Training Hyperparameters}
\label{app:training_hyperparams}

\begin{table}[htbp]
\centering
\small
\caption{Training hyperparameters used in our experiments.}
\label{tab:training_hyperparams}
\begin{tabular}{lc}
\toprule
\textbf{Hyperparameter} & \textbf{Value} \\
\midrule
Backbone models & Qwen2.5-3B-Instruct / Qwen2.5-1.5B-Instruct / Qwen3-0.6B \\
Training examples & 11,895 \\
Validation examples & 512 \\
Test examples & 512 \\
Training steps & 400 \\
Training batch size & 32 \\
Number of sampled responses $G$ & 8\\
Diversity coefficient $\lambda$ & 0.0, 0.2, 0.4, 0.6, 0.8, 1.0 \\
KL coefficient $\beta$ & 0.001 \\
Actor KL loss type & low-var KL \\
Clipping parameter $\epsilon$ & 0.2 \\
Learning rate & $1\times 10^{-6}$ \\
Learning-rate warmup steps & 10 \\
Batch size & 32 \\
Optimizer & AdamW \\
Weight decay & 0.01 \\
Gradient clipping & 0.9 \\
Max prompt length & 4096 \\
Max CharacterEval prompt length & 4096 \\
Max response length & 512 \\
Temperature & 1.0 \\
Top-$p$ & 1.0 \\
Embedding model for diversity & Qwen3-Embedding-0.6B \\
Multi-turn reward model & BaichuanCharRM \\
CharacterEval score range & $[1,5]$ \\
\bottomrule
\end{tabular}
\end{table}
\end{document}